\documentclass[10pt,twocolumn,letterpaper]{article}
\usepackage[pagenumbers]{cvpr} 
\usepackage[accsupp]{axessibility}
\definecolor{cvprblue}{rgb}{0.21,0.49,0.74}
\usepackage[pagebackref,breaklinks,colorlinks,allcolors=cvprblue]{hyperref}

\usepackage{multirow}
\usepackage{multicol}
\usepackage{pifont}
\usepackage{rotating}
\usepackage{colortbl}
\usepackage{xcolor}
\usepackage{adjustbox}
\usepackage{pifont}
\usepackage{seqsplit}

\def\paperID{15576} 
\def\confName{CVPR}
\def\confYear{2025}

\title{DropGaussian: Structural Regularization for Sparse-view Gaussian Splatting}

\author{
    Hyunwoo Park \quad Gun Ryu \quad Wonjun Kim\thanks{Corresponding author} \\ 
    Konkuk University\\
    {\tt\small \{pzls, fbrjs15, wonjkim\}@konkuk.ac.kr}
}

\begin{document}
\maketitle
\begin{abstract}

Recently, 3D Gaussian splatting (3DGS) has gained considerable attentions in the field of novel view synthesis due to its fast performance while yielding the excellent image quality.
However, 3DGS in sparse-view settings (e.g., three-view inputs) often faces with the problem of overfitting to training views, which significantly drops the visual quality of novel view images.
Many existing approaches have tackled this issue by using strong priors, such as 2D generative contextual information and external depth signals.
In contrast, this paper introduces a prior-free method, so-called DropGaussian, with simple changes in 3D Gaussian splatting.
Specifically, we randomly remove Gaussians during the training process in a similar way of dropout, which allows non-excluded Gaussians to have larger gradients while improving their visibility.
This makes the remaining Gaussians to contribute more to the optimization process for rendering with sparse input views.
Such simple operation effectively alleviates the overfitting problem and enhances the quality of novel view synthesis.
By simply applying DropGaussian to the original 3DGS framework, we can achieve the competitive performance with existing prior-based 3DGS methods in sparse-view settings of benchmark datasets without any additional complexity.
The code and model are publicly available at:  
\texttt{\href{https://github.com/DCVL-3D/DropGaussian_release}{\textcolor{magenta}{\seqsplit{https://github.com/DCVL-3D/DropGaussian\textunderscore release}}}}.

\end{abstract}
\vspace{-3mm}
\section{Introduction}
\label{sec:intro}
As the demand for realistic renderings and their applications increases rapidly, novel view synthesis (NVS) has become an essential technique. 
Recently, the neural radiance field (NeRF)~\cite{mildenhall20NeRF} has been introduced in literature, which has a good ability to encode a given 3D scene into the implicit radiance field through learnable parameters of the neural network.
Even though NeRF and its diverse variants have shown the remarkable performance, most previous methods require time-consuming processes for rendering as well as training.
Meanwhile, a new technique, so-called 3D Gaussian splatting (3DGS)~\cite{kerbl233DGS}, has emerged in 2023 and is becoming a mainstream in the field of NVS.
3DGS is a method that generates an explicit radiance field, composed of a set of 3D Gaussians positioned in 3D space.
Based on the property of point-based explicit representations, 3DGS demonstrates real-time operations for rendering while maintaining the rendering quality for novel view inputs. 
However, with sparse input views, 3DGS is still prone to overfitting due to insufficient cues for visual appearance and geometric layout in understanding a given 3D scene.

\begin{figure}
\centering
\centerline{\includegraphics[width=1.0\columnwidth]{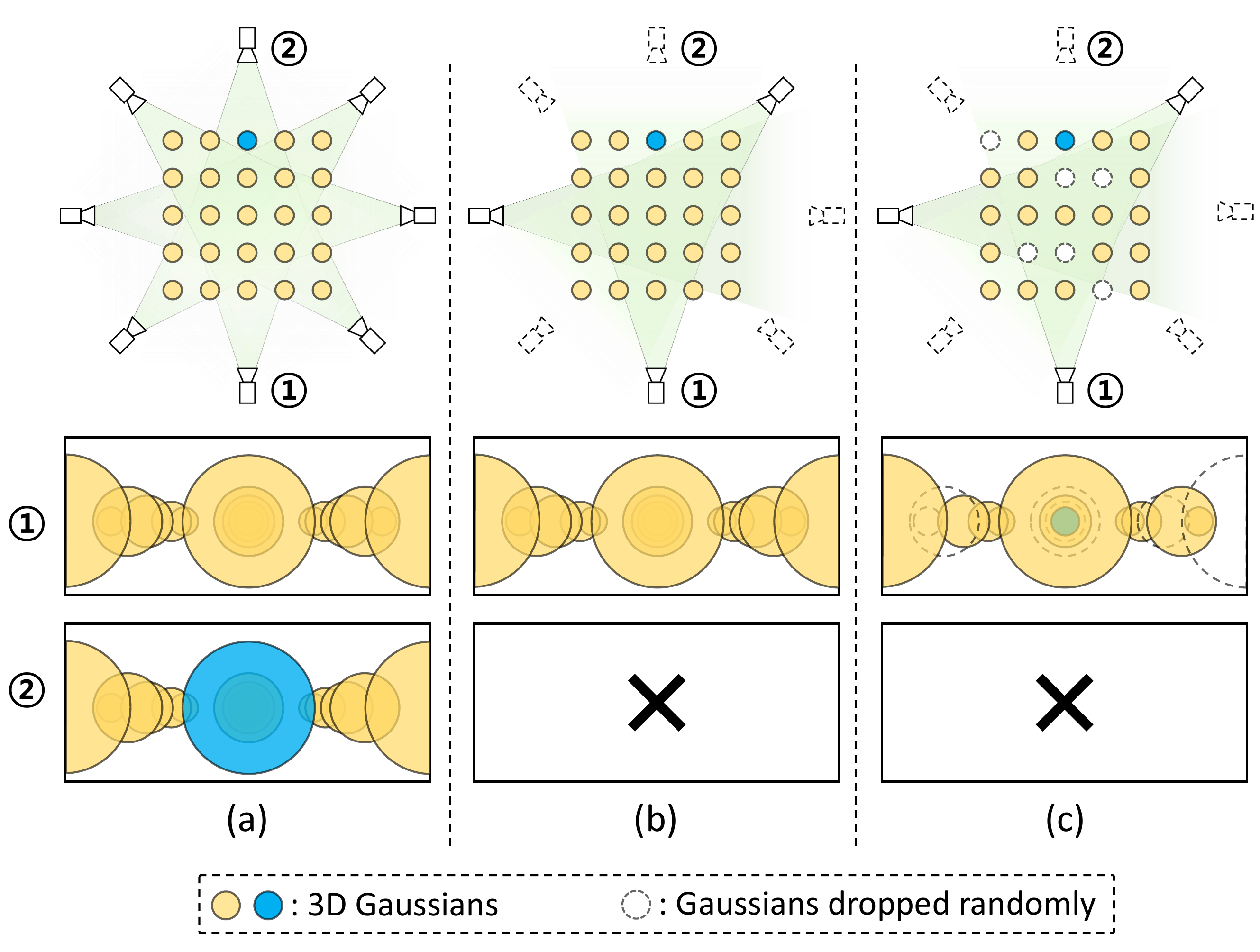}}
\vspace{-2mm}
\caption{\label{fig:1}
(a) Traditional settings of 3DGS. (b) Sparse-view set tings of 3DGS. (c) Effect of DropGaussian in sparse-view settings. The rendered outputs at each viewpoint are visualized in (a), (b), and (c), respectively.}
\vspace{-6mm}
\end{figure}
Despite substantial efforts in the field of 3DGS, the optimization of 3D Gaussians only with sparse input views still remains challenging. 
To address this challenge, several approaches have begun to leverage the prior information of a given scene. 
For example, there have been meaningful attempts to adopt the result of monocular depth estimation as an external supervisory signal for imposing Gaussians on appropriate positions~\cite{chung24, zhu24FSGS, li24Dngaussian}.
However, without considering the world coordinate, estimated depth scales vary across different views, which makes consistent regularization difficult.
Even though the 2D generative contextual information also has been successfully employed to guide the rendering process to yield more realistic results~\cite{xiong23sparsegs}, it requires high computational costs and often leads to unstable optimization due to the stochastic sampling process.
On the other hand, the optical flow has been utilized to regularize the pixel-wise correspondence of 3D Gaussians between 3D Gaussians in sparse-view conditions~\cite{paliwal24coherentgs}. 
Although such prior-based approaches have been actively explored to mitigate the overfitting problem driven by sparse input views, most of them have respective limitations such as error propagation and high computational burden.

In this paper, unlike previous methods relying on strong priors, we present a prior-free method that requires only a simple modification of 3DGS without additional computational costs.
In conventional settings of 3DGS, Gaussians, which are far from the camera and probably occluded in a specific viewpoint, can be visible in other viewpoints as shown in Fig.~\ref{fig:1} (a).
In contrast, such Gaussians are often excluded from the field of view in sparse-view settings, resulting in receiving less gradient feedback due to the low visibility caused by occlusion between other Gaussians (see Fig.~\ref{fig:1}(b)).
This ultimately leads to overfitting to a small number of training views.
The key idea of the proposed method is to randomly remove Gaussians, so-called \textit{DropGaussian}, during training instead of adopting the prior information for sparse input views.
Based on our DropGaussian scheme, the remaining Gaussians are provided with the opportunity to be more visible as illustrated in Fig.~\ref{fig:1} (c).
Such simple operation makes the optimization process with sparse input views be more balanced in the sense that even less visible Gaussians receive adequate attentions during training (see Fig.~\ref{fig:1}(c)).
Consequently, the model is able to figure out the whole layout of a given 3D scene in a comprehensive manner.
This effectively generalizes the rendering performance to novel views and mitigates the risk of overfitting to the limited number of training views.
In addition, we have observed that overfitting primarily manifests during the later stage of training rather than the initial phase under sparse-view conditions.
Based on this observation, we further proposed to progressively apply our DropGaussian scheme to the training process of 3DGS.
The main contribution of the proposed method can be summarized as follows:

\begin{itemize}[leftmargin=4mm]
\item We propose a simple yet powerful regularization technique, so-called DropGaussian, for rendering with sparse input views.
By randomly eliminating Gaussians during the training process, DropGaussian gives the opportunity for the remaining Gaussians to be more visible with larger gradients, which make them to meaningfully contribute to the optimization process of 3DGS.
This is fairly desirable to alleviate the overfitting problem occurring in sparse-view conditions. 
\item Through various experiments on benchmark datasets, we identified that overfitting predominantly occurs during the later stage of training rather than the initial phase with sparse input views.
Based on this observation, we propose to progressively increase the ratio of dropping Gaussians during the training process.
This adaptive strategy effectively mitigates the overfitting problem without unexpected impact on the rendering performance at the initial phase.
\end{itemize}

\section{Related Works}
In this Section, we provide a brief review of recent methods in novel view synthesis and explore various approaches specifically designed to address challenges related to sparse input views.
\subsection{Novel View Synthesis}
Recently, the neural radiance field (NeRF)~\cite{mildenhall20NeRF} has achieved the significant progress in encoding a given 3D scene into implicit radiance fields by mapping 3D coordinates and the view direction to color and density values through a simple neural network.
Numerous variants have been introduced particularly focusing on improving the rendering quality ~\cite{barron21mip,barron22mip360,verbin22refnerf}, accelerating the training speed~\cite{fridovich22plenoxels,muller22instant}, and extending the model to handle more complex scenarios such as dynamic scenes~\cite{park21hypernerf} or unconstrained scenes~\cite{martin21nerfw}.
Despite such extensive efforts on enhancing the capability of NeRF, most previous methods require time-consuming processes for both training and rendering, which limits their practical applicability.
To address this limitation, a new approach, which is known as 3D Gaussian splatting (3DGS)~\cite{kerbl233DGS}, has been emerged.
3DGS represents a given scene by utilizing a set of 3D Gaussians and renders images via a differentiable rasterization scheme.
By replacing the neural network with point-based 3D Gaussians, 3DGS remarkably enhances the efficiency of training and rendering, enabling real-time applications of 3DGS. 
Inspired by plentiful possibilities of 3DGS, various follow-up studies have been introduced in most recent days, e.g., improving the rendering quality~\cite{yu24mipsplat}, addressing the memory inefficiency~\cite{niedermayr24compressGS, lee24compactGS}, and considering the temporal relationship for dynamic scenes~\cite{wu244DGS}.
\begin{figure*}[t]
\centerline{\includegraphics[width=1\textwidth]{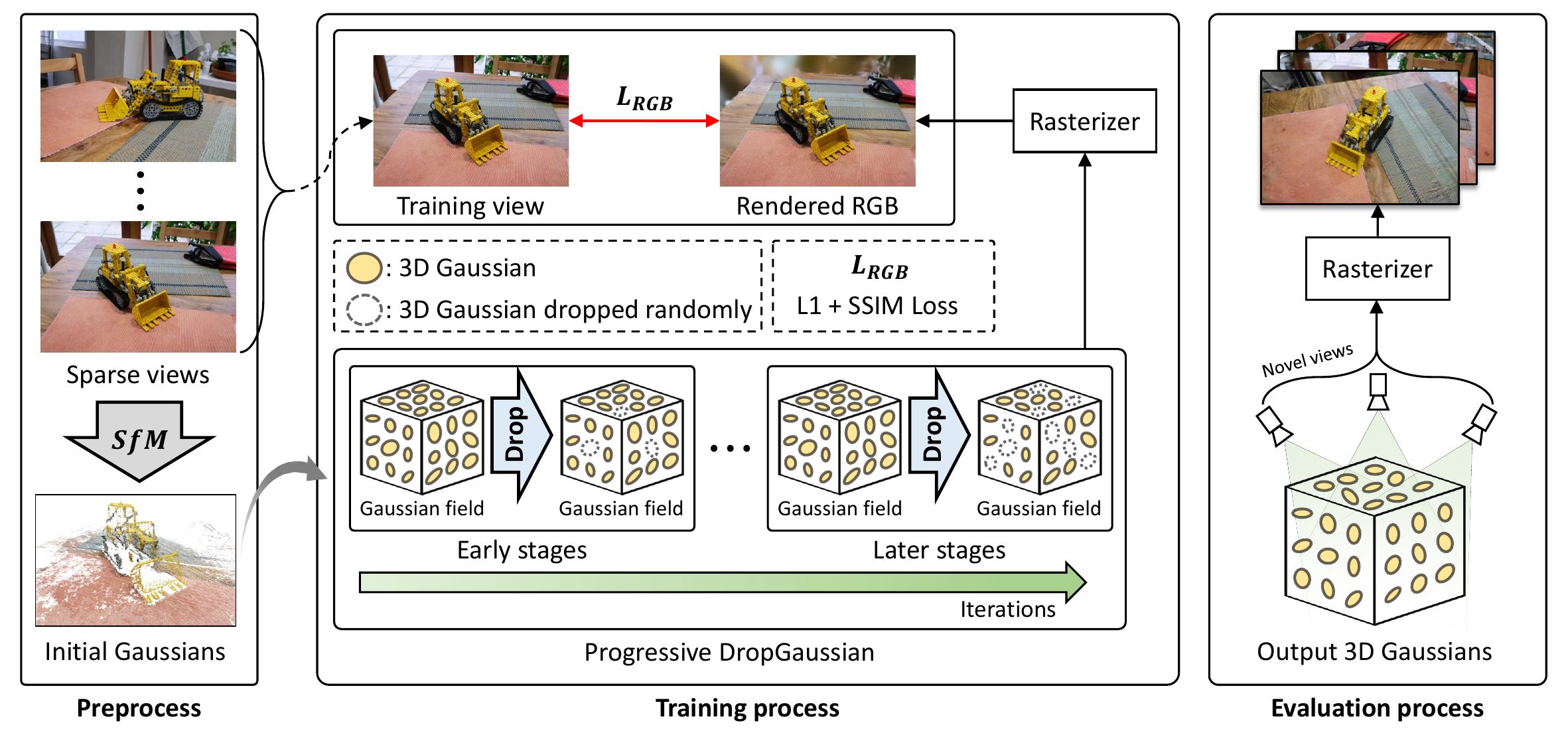}}
\vspace{-2mm}
\caption{\label{fig:concept} The overall framework of the proposed method for 3DGS in sparse-view settings. Our DropGaussian scheme improves the visibility of Gaussians even far from the camera by randomly dropping Gaussians during the training process, thereby mitigating overfitting to the limited number of training views. In contrast, during the test phase, all the Gaussians are rendered to generate high-quality RGB images, ensuring that the complete scene representation is utilized for novel view synthesis.
}
\vspace{-5mm}
\end{figure*}
\subsection{Novel View Synthesis with Sparse-Views}
Even though NeRF and 3DGS have shown surprising rendering performance, a large number of input images are required to guarantee the high-quality result.
However, in real-world scenarios, only a small number of images are available (i.e., sparse input views), which leads to the performance degradation by the overfitting problem.
Before the emergence of 3DGS, NeRF-based methods have attempted to resolve this problem in various ways.
In the early stages, Yu \textit{et al.}~\cite{yu21pixelnerf} employed a pre-trained CNN encoder to incorporate the contextual information into NeRF through transfer learning.
Jain \textit{et al.}~\cite{jain21dietnerf} designed a semantic consistency loss by utilizing CLIP embeddings~\cite{radford21CLIP} to effectively handle unseen views.
Furthermore, Niemeyer \textit{et al.}~\cite{niemeyer2022regnerf} introduced a patch-based color and depth regularization method, making geometry and appearance across neighboring regions be consistent.
Most recently, Yang \textit{et al.}~\cite{yang23freenerf} suggested a simple yet powerful scheme, i.e., frequency regularization, which improves the generalization ability by adjusting the input frequency range of NeRF.
Wang \textit{et al.}~\cite{wang23sparsenerf} introduced a local depth ranking constraint for sparse input views, ensuring that the estimated depth order within local regions remains consistent with coarse observations, thereby mitigating inaccuracies in sparse scenarios.

On the other hand, several methods in 3DGS have begun to be actively studied for sparse input views.
Specifically, Chung \textit{et al.}~\cite{chung24} attempted to clarify the depth ambiguity by injecting the learnable parameter during training, which often occurs between different views.
In a similar way, Li \textit{et al.}~\cite{li24Dngaussian} proposed a global-local depth normalization technique, which normalizes depth values on patch-based local scales, thereby accurately refocusing on small changes of the local depth.
Instead of conducting depth-based regularization, Zhu \textit{et al.}~\cite{zhu24FSGS} introduced a Gaussian unpooling scheme, which is designed to generate new Gaussians by leveraging graph-based proximity scores from the nearest \textit{K} neighbors. 
Xiong \textit{et al.}~\cite{xiong23sparsegs} proposed to apply score distillation sampling loss~\cite{poole22Dreamfusion} to the training process for refining plausible details in regions with limited coverage in training views (i.e., sparse input views) and generating more complete 3D representations. 
In particular, Paliwal \textit{et al.}~\cite{paliwal24coherentgs} utilized a pre-trained optical flow model~\cite{shi23flowformer++} to regularize the pixel-wise correspondence between 3D Gaussians, which efficiently alleviates the overfitting problem by sparse input views.

\vspace{-3mm}
\section{Proposed Method}
\vspace{-2mm}
The proposed method aims to improve the updating process of Gaussian parameters in sparse-view settings.
In this Section, we introduce DropGaussian, which randomly removes Gaussians during training, thereby increasing the updating opportunities for the remaining ones under sparse-view conditions thereby improving the visibility for the remaining Gaussians under sparse-view conditions.
It is noteworthy that the ratio of dropping Gaussians progressively increases, which better mitigates the overfitting problem in the later stage of training.
The overall framework of the proposed method is shown in Fig.~\ref{fig:concept}.
\vspace{-2mm}
\subsection{Preliminaries}
3D Gaussian splatting (3DGS)~\cite{kerbl233DGS} explicitly represents a given scene by using a set of point-based 3D Gaussians. 
Each Gaussian is defined by a center \(\mu\in{\mathbb{R}^3}\), a scaling factor \(s\in{\mathbb{R}^3}\), and a rotation quaternion \(q\in{\mathbb{R}^4}\).
The basis function for the \(i\)-th Gaussian, i.e., \(G_i\), is given by:
\begin{equation}
G_i(\boldsymbol{x}) = \exp\left( -\frac{1}{2} (\boldsymbol{x} - \boldsymbol{\mu}_i)^T \boldsymbol{\Sigma}_i^{-1} (\boldsymbol{x} - \boldsymbol{\mu}_i) \right),
\label{eq:1}
\end{equation}
where the covariance matrix \(\Sigma_i\) is approximated by the combination of the scaling factor \(s_i\) and the rotation quaternion \(q_i\), which determines size and orientation of the Gaussian in the 3D space.
In addition to these geometric attributes, each Gaussian also possesses an opacity value \(o_i\in\mathbb{R}\) and a \textit{K}-dimensional color feature \(f_i\in\mathbb{R}^K\).
The color feature is typically represented by using spherical harmonic (SH) coefficients, which efficiently encode the lighting effect according to different directions.
For rendering, the color of the \(i\)-th Gaussian, i.e., RGB value, is computed from these SH coefficients.
The final color \( C(p) \) at a given pixel \( p \) is computed by accumulating the weighted color contributions of all Gaussians, which can be formulated as follows:
\begin{equation} C(p) = \sum_{i=1}^{N} c_i \alpha_i T_i, \label{eq:2
} \end{equation}
where \( c_i \) is the RGB value derived from the color feature \(f_i\) of the \( i \)-th Gaussian.
\( \alpha_i \) denotes the product of the projected 2D Gaussian and the learned opacity value \(o_i\) of the \(i\)-th Gaussian.
\(T_i\) represents the accumulated transparency along the view direction of the \(i\)-th Gaussian.
\( N \) denotes the total number of Gaussians contributing to the final color of the pixel \(p\).
By following~\cite{kerbl233DGS}, \( T_i \) and \( \alpha_i \) can be computed as follows:
\begin{equation} \label{eq:3}
T_i = \prod_{j=1}^{i-1} (1 - \alpha_j), \quad\alpha_i = o_i\cdot g_i^{2D},
\end{equation}
where \(g_i^{2D}\) is the 2D Gaussian, which is obtained by projecting the \(i\)-th 3D Gaussian onto the image plane.

\subsection{DropGaussian}
In sparse-view conditions, the transmittance \(T_i\) of 3D Gaussians, which are far from the camera and thus have a high probability of being occluded by other Gaussians, becomes relatively low as illustrated in Fig.~\ref{fig:1}(b).
Since the visible range of such Gaussians is strictly limited due to a small number of input views, Gaussian attributes, e.g., scale, color, opacity, etc., are not able to be actively updated, leading to reduction of their contributions to the overall rendering process.
Consequently, the gradient feedback of the optimization process in 3DGS is not sufficiently provided to the corresponding Gaussians, and it eventually results in overfitting to a small number of training views (i.e., sparse input views).
\begin{figure}
\centering
\centerline{\includegraphics[width=1.0\columnwidth]{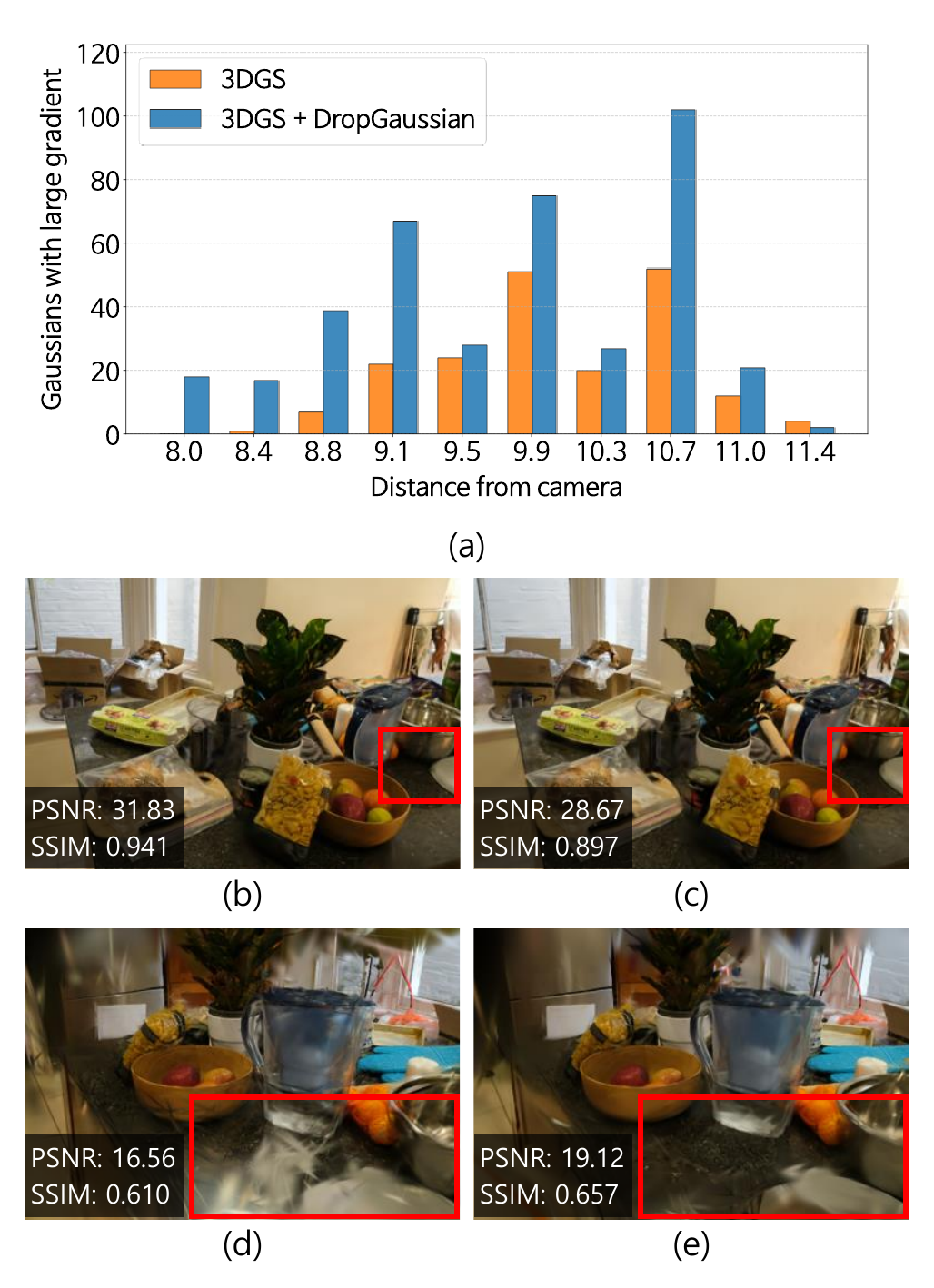}}
\vspace{-3mm}
\caption{\label{fig:3}
(a) Distribution of Gaussians having larger gradients according to the distance from camera. Note that the y-axis denotes the number of Gaussians with the gradient value, which is larger than the threshold value of densification in 3DGS. (b) Rendering results for the training view by the 3DGS. (c) Rendering results for the training view by the 3DGS with our DropGaussian scheme. (d) Rendering result for the novel view by 3DGS. (e) Rendering result for the novel view by 3DGS with our DropGaussian scheme.}
\vspace{-5mm}
\end{figure}

To address this problem, we propose DropGaussian, a structural regularization technique which randomly removes a set of Gaussians during the training process. 
Specifically, the dropping rate \(r\) is firstly defined, for example, \(r\) is set to 0.1 for \(10\%\) removal of total Gaussians.
Since the cumulative opacity contributing to the color value of each pixel is probably decreased by this dropping process, we propose to apply the compensation factor to the opacity value of the remaining Gaussians as follows:
\begin{equation} \label{eq:5}
\tilde{o}_i = M(i)\cdot o_i,
\end{equation}
where \(M(i)\) indicates the compensation factor for the \(i\)-th Gaussian, which assigns \(\frac{1}{(1-r)}\) to the remaining Gaussians and 0 otherwise.
It is noteworthy that the total contribution of Gaussians to the color value of each pixel is successfully maintained by the scaling effect of the compensation factor even with the dropping process.
Moreover, Gaussians that are far from the camera can have large gradients by DropGaussian because their visibility is efficiently improved as shown in Fig.~\ref{fig:1} (c).

To show the effect of our DropGaussian scheme, we further analyze the distribution of Gaussians with their gradient values according to the distance from the camera.
As shown in Fig.~\ref{fig:3} (a), Gaussians can have larger gradients by DropGaussian even though they are somewhat far from the camera.
Note that we only count Gaussians with gradients whose value is larger than the threshold value for densification in 3DGS (i.e., 0.0005 in this example).
This makes 3DGS to be robust to the overfitting problem when rendering novel view images under sparse-view conditions (see Fig.~\ref{fig:3} (e)).

\begin{figure}
\centering
\centerline{\includegraphics[width=1.0\columnwidth]{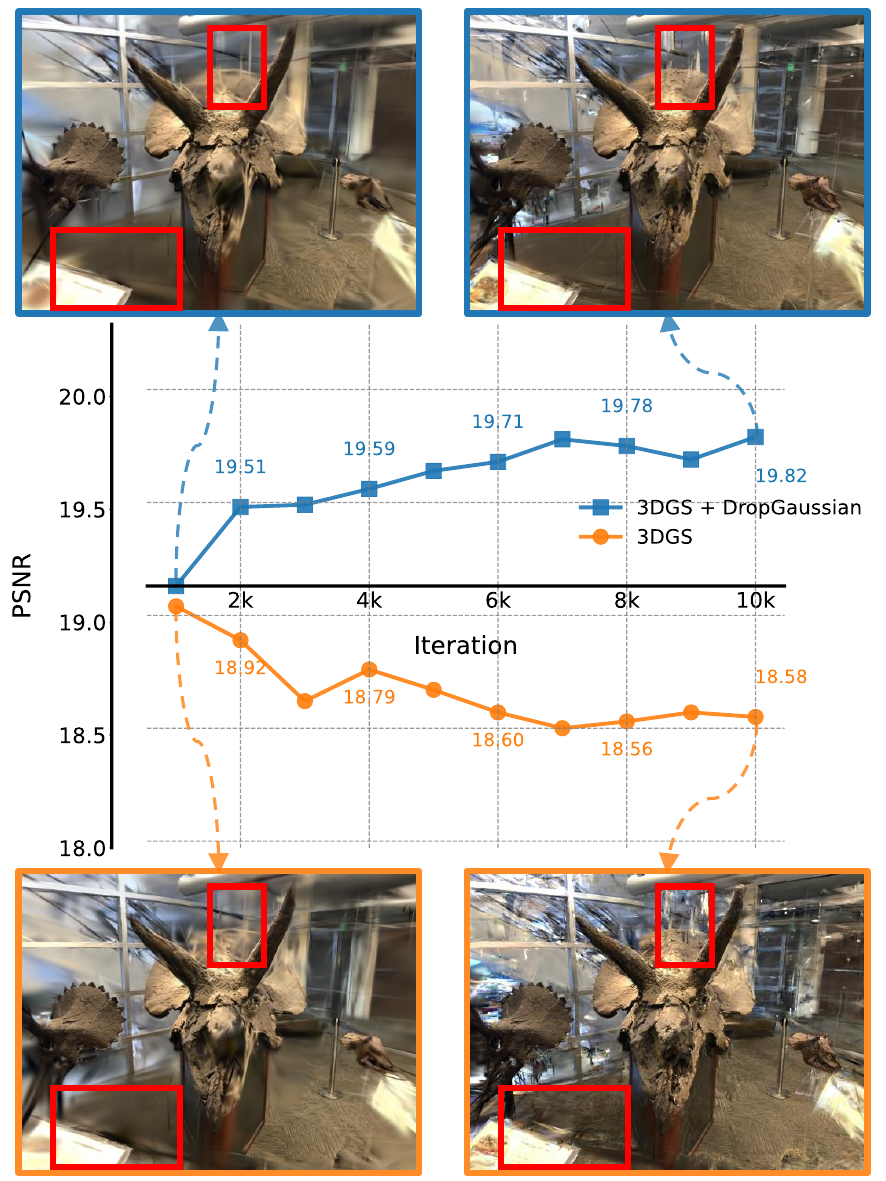}}
\caption{\label{fig:4}
Changes of PSNR values during the training process on the LLFF dataset for 3DGS and 3DGS with DropGaussian. Note that rendering results for novel views at 1,000 and 10,000 iterations are also given, highlighting the intensification of overfitting in later training stages and the effectiveness of DropGaussian in mitigating this issue. The red boxes indicate regions where overfitting occurs.}
\vspace{-5mm}
\end{figure}

Furthermore, we have observed that the tendency for overfitting becomes stronger under sparse-view conditions as the training process progresses as shown in Fig.~\ref{fig:4}.
To resolve this problem, we propose to adjust the dropping rate according to the current iteration index \(t\) as follows:
\begin{equation}
r_t=\gamma\cdot\frac{t}{t_{\mathrm{total}}},
\end{equation}
where \(\gamma\in\{0,1\}\) denotes the scaling factor for the dropping rate.
\(t_{\mathrm{total}}\) is the total iterations of the training process.
This progressive adjustment strategy strengthens the regularization effect as the training process progresses.

\subsection{Loss Function}
The proposed method is trained based on the traditional color reconstruction loss.
Following 3D Gaussian splatting, the color reconstruction loss is composed of L1 loss and D-SSIM loss, which measures the structural similarity between the rendered image \( \hat{I} \) and the ground-truth image \( I \), given as~\cite{kerbl233DGS}:
\begin{equation}
\mathcal{L}_{\text{color}} = \mathcal{L}_1(\hat{I}, I) + \lambda \mathcal{L}_{\text{D-SSIM}}(\hat{I}, I),
\end{equation}
where \(\lambda\) denotes the weighting factor that makes a balance between the contribution of \(\mathcal{L}_1\) and \(\mathcal{L}_{\text{D-SSIM}}\), which is set to 0.2.

\section{Experimental Results}
\begin{figure*}[t]
\centerline{\includegraphics[width=1.0\textwidth]{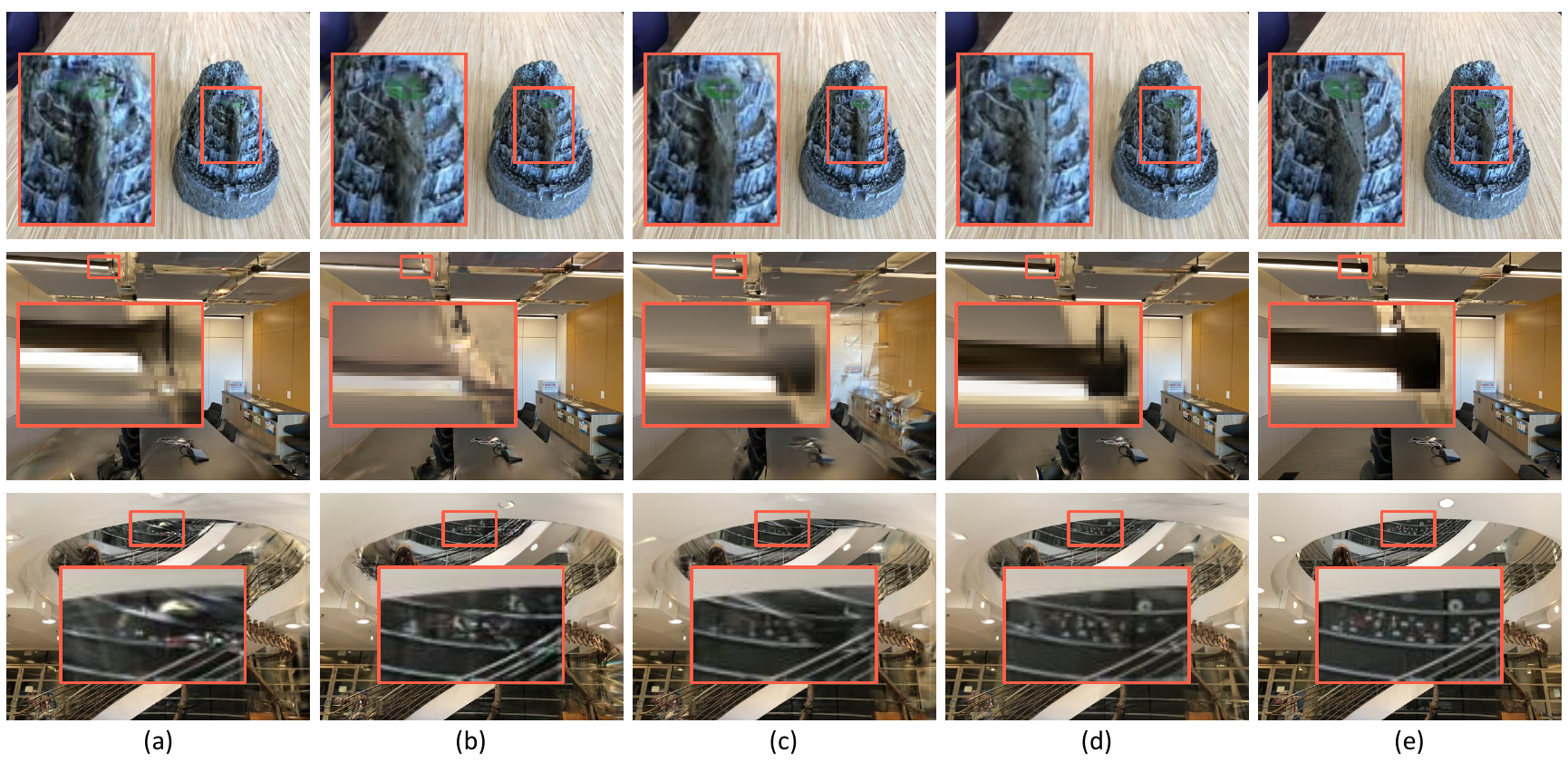}}
\vspace{-3mm}
\caption{\label{fig:6} Results of novel view rendering on the LLFF~\cite{barron22mip360} dataset. (a) Results by 3DGS~\cite{kerbl233DGS}. (b) Results by FSGS~\cite{zhu24FSGS}. (c) Results by CoR-GS~\cite{zhang24CoR-GS}. (d) Results by the proposed method. (e) GT image.
}
\vspace{-2mm}
\end{figure*}
\begin{table*}[h!]
\scriptsize
\centering
\adjustbox{max width=\textwidth}{%
\begin{tabular}{c l| c c c | c c c | c c c}
\toprule
\multicolumn{2}{c}{\multirow{2}{*}[-0.5ex]{Methods}} & \multicolumn{3}{c}{3-view} & \multicolumn{3}{c}{6-view} & \multicolumn{3}{c}{9-view}\\ \cmidrule(lr){3-5}\cmidrule(lr){6-8}\cmidrule(lr){9-11}
\multicolumn{2}{c}{} & \multicolumn{1}{c}{PSNR($\uparrow$)} & \multicolumn{1}{c}{SSIM($\uparrow$)} & \multicolumn{1}{c}{LPIPS ($\downarrow$)} & \multicolumn{1}{c}{PSNR($\uparrow$)} & \multicolumn{1}{c}{SSIM($\uparrow$)} & \multicolumn{1}{c}{LPIPS ($\downarrow$)} & \multicolumn{1}{c}{PSNR($\uparrow$)} & \multicolumn{1}{c}{SSIM($\uparrow$)} & \multicolumn{1}{c}{LPIPS ($\downarrow$)}\\ \midrule
\toprule
\multirow{5}{*}[0ex]{\centering\rotatebox{90}{NeRF-based}} 
& Mip-NeRF~\cite{barron21mip} & 16.11 & 0.401 & 0.460 & 22.91 & 0.756 & 0.213 & 24.88 & 0.826 & 0.170\\
& DietNeRF~\cite{jain21dietnerf} & 14.94 & 0.370 & 0.496 & 21.75 & 0.717 &  0.248 & 24.28 &  0.801 & 0.183\\ 
& RegNeRF~\cite{niemeyer2022regnerf} & 19.08 &  0.587 & 0.336 & 23.10 & 0.760 & 0.206 & 24.86 &  0.820 &  0.161\\
& FreeNeRF~\cite{yang23freenerf} & 19.63 & 0.612 & 0.308 & 23.73 &  0.779 & 0.195  & 25.13 &  0.827 & 0.160\\
& SparseNeRF~\cite{wang23sparsenerf} & 19.86 & 0.624 & 0.328 & - &  - & - & - & - &  -\\
\midrule
\multirow{5}{*}[0ex]{\centering\rotatebox{90}{3DGS-based}} 
& 3DGS~\cite{kerbl233DGS} & 19.22 & 0.649 & 0.229\cellcolor{yellow!25} & 23.80 & 0.814 & 0.125\cellcolor{yellow!25} & 25.44 \cellcolor{yellow!25}& 0.860 \cellcolor{yellow!25}& 0.096\cellcolor{yellow!25}\\
& DNGaussian~\cite{li24Dngaussian} & 19.12 &  0.591 &  0.294 & 22.18 & 0.755 &  0.198 & 23.17 & 0.788 & 0.180\\
& FSGS~\cite{zhu24FSGS} & 20.43\cellcolor{yellow!25} & 0.682 \cellcolor{yellow!25}& 0.248 & 24.09 \cellcolor{yellow!25}& 0.823\cellcolor{yellow!25} & 0.145 & 25.31 & 0.860 & 0.122\\

& CoR-GS~\cite{zhang24CoR-GS} & 20.45\cellcolor{orange!25} & 0.712\cellcolor{orange!25} & \cellcolor{red!25}0.196 & 24.49 \cellcolor{orange!25}& \cellcolor{red!25}0.837 & 0.115\cellcolor{red!25} & 26.06 \cellcolor{orange!25}& 0.874\cellcolor{red!25} & 0.089\cellcolor{orange!25} \\

&Ours & \cellcolor{red!25}20.76 & 0.713 \cellcolor{red!25}& 0.200\cellcolor{orange!25} & 24.74\cellcolor{red!25} & 0.837\cellcolor{red!25} & 0.117\cellcolor{orange!25} & 26.21\cellcolor{red!25}& 0.874\cellcolor{red!25} & 0.088\cellcolor{red!25}\\ \bottomrule

\end{tabular}}
\vspace{-2mm}
\caption{\label{table:llff} Performance comparisons of sparse-view synthesis on LLFF dataset~\cite{mildenhall19llff}. The best, second-best, and third-best entries are marked in red, orange, and yellow, respectively.}
\vspace{-5mm}
\end{table*}

\subsection{Training}
All experiments were conducted by using the PyTorch framework~\cite{Paszke17torch}, running on an Intel E5-1650 v4@3.60GHz CPU and an NVIDIA RTX 3090Ti GPU.
We employed the Adam optimizer~\cite{Kingma15adam} to train all model parameters, with momentum factors set to 0.9 and 0.999, respectively.
The proposed method is trained for 10,000 iterations with densification performed every 100 iterations.
The gradient threshold of densification is set to \(5\times 10^{-4}\) as used in~\cite{zhu24FSGS}.
\subsection{Datasets and Evaluation Metrics}
\noindent\textbf{Dataset.} For performance evaluation of the proposed method, three representative benchmarks, i.e., LLFF~\cite{mildenhall19llff}, Mip-NeRF360~\cite{barron22mip360}, and Blender~\cite{mildenhall20NeRF}, are employed.
We follow the settings used in previous works with the same split of LLFF, Mip-NeRF360, and blender datasets, which consist of 3, 12, and 8 training views, respectively.
The downsampling rates are set to 8 for both LLFF and Mip-NeRF360 while 2 is used for Blender.

\noindent\textbf{Evaluation metrics.}
For the quantitative evaluation, we use three metrics, i.e., peak signal-to-noise ratio (PSNR), structural similarity index (SSIM)~\cite{wang04ssim}, and learned perceptual image patch similarity (LPIPS) ~\cite{zhang18lpips}, which have been generally adopted in this field.
Specifically, PSNR measures the average peak error between the rendered image and the ground truth. 
SSIM computes the structural similarity based on luminance, contrast, and texture information. 
On the other hand, LPIPS computes the perceptual distance by utilizing learned features, which is useful to figure out underlying differences that are not reflected by traditional metrics.

\begin{figure*}[t]
\centerline{\includegraphics[width=1.0\textwidth]{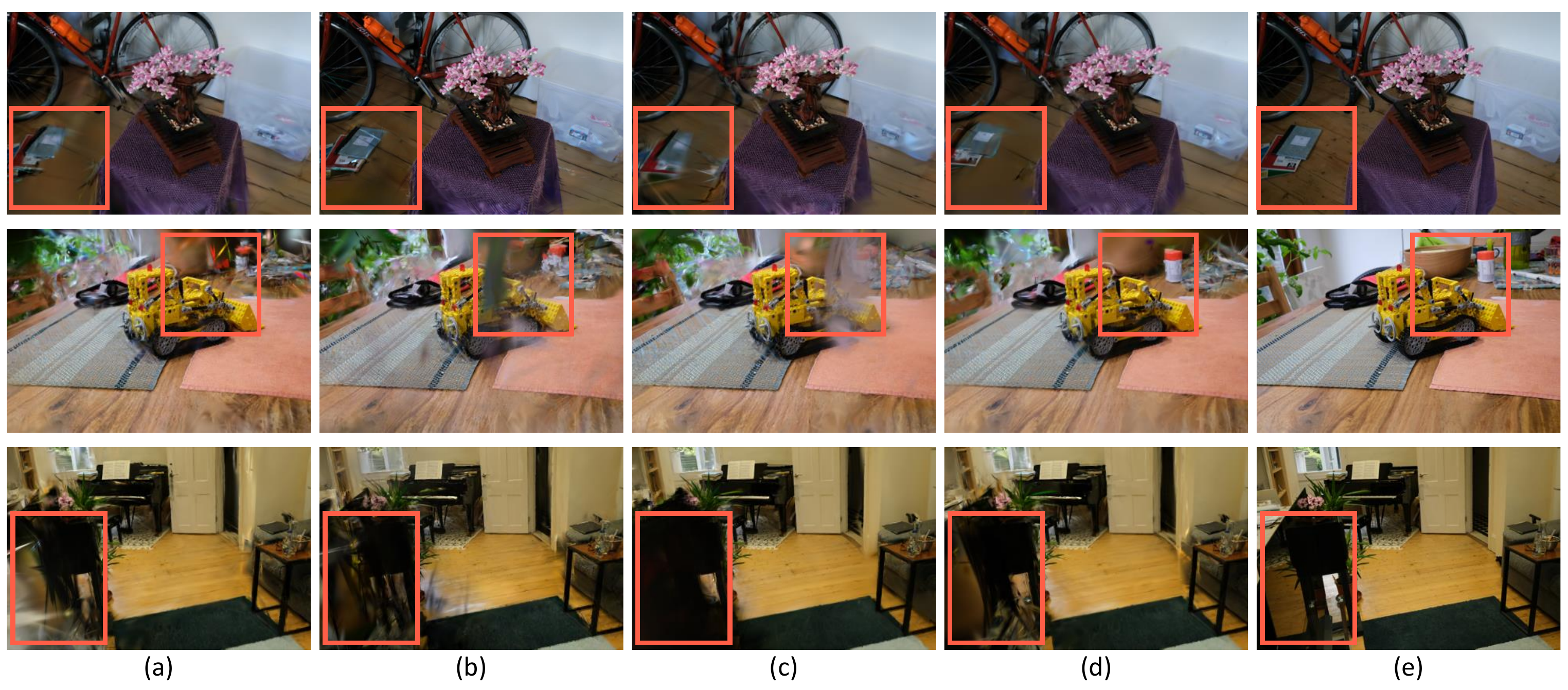}}
\vspace{-3mm}
\caption{\label{fig:5} Results of novel view rendering on the Mip-NeRF360~\cite{barron22mip360} dataset. (a) Results by 3DGS~\cite{kerbl233DGS}. (b) Results by FSGS~\cite{zhu24FSGS}. (c) Results by CoR-GS~\cite{zhang24CoR-GS}. (d) Results by the proposed method. (e) GT image.
}
\vspace{-2mm}
\end{figure*}

\subsection{Performance Evaluation}
\vspace{-1mm}
\noindent\textbf{Quantitative evaluation.}
To demonstrate the effectiveness of the proposed method in addressing overfitting, we compare it with previous methods in rendering with sparse input views, i.e., MipNeRF~\cite{barron21mip}, DietNeRF~\cite{jain21dietnerf}, RegNeRF~\cite{niemeyer2022regnerf}, FreeNeRF~\cite{yang23freenerf}, SparseNeRF~\cite{wang23sparsenerf}, DNGaussian~\cite{li24Dngaussian}, FSGS~\cite{zhu24FSGS}, and CoR-GS~\cite{zhang24CoR-GS}.
First of all, the performance comparison on the LLFF dataset is shown in Table~\ref{table:llff}.
As can be seen, the proposed method shows the meaningful improvement for the rendering performance with a very simple operation. Specifically, the proposed method achieves the highest PSNR of 20.76 in the 3-view setting, while surpassing all NeRF-based and 3DGS-based methods. For both 6-view and 9-view settings, our approach still achieves the competitive performance compared to the state-of-the-art methods (e.g., CoR-GS~\cite{zhang24CoR-GS}) without any increase in the computational complexity.
\begin{table*}[h!]
\scriptsize
\centering
\adjustbox{max width=\textwidth}{%
\centering
\renewcommand{\arraystretch}{1.1} 
\begin{tabular}{c l|  c c c | c c c} 
\toprule
\multicolumn{2}{c}{\multirow{2}{*}[-0.5ex]{Methods}} & \multicolumn{3}{c}{12-view} & \multicolumn{3}{c}{24-view}\\ \cmidrule(lr){3-5}\cmidrule(lr){6-8}
\multicolumn{2}{c}{} & \multicolumn{1}{c}{PSNR($\uparrow$)} & \multicolumn{1}{c}{SSIM($\uparrow$)} & \multicolumn{1}{c}{LPIPS ($\downarrow$)} & \multicolumn{1}{c}{PSNR($\uparrow$)} & \multicolumn{1}{c}{SSIM($\uparrow$)} & \multicolumn{1}{c}{LPIPS ($\downarrow$)}\\ \midrule
& 3DGS~\cite{kerbl233DGS} & 18.52 & 0.523 & 0.415\cellcolor{orange!25} & 22.80 & 0.708 & 0.276\\
& FSGS~\cite{zhu24FSGS} & 18.80\cellcolor{yellow!25}&  0.531\cellcolor{yellow!25} & 0.418 & 23.70 \cellcolor{orange!25}& 0.745\cellcolor{orange!25} & 0.230\cellcolor{orange!25}\\
& CoR-GS~\cite{zhang24CoR-GS} & 19.52\cellcolor{orange!25} & 0.558\cellcolor{orange!25} & 0.418 \cellcolor{yellow!25}& 23.39 \cellcolor{yellow!25}& 0.727\cellcolor{yellow!25} & 0.271\cellcolor{yellow!25} \\
&Ours & 19.74\cellcolor{red!25} & 0.577\cellcolor{red!25} & 0.364\cellcolor{red!25} & 24.13\cellcolor{red!25} & 0.762\cellcolor{red!25} & 0.225\cellcolor{red!25}\\ \bottomrule
\end{tabular}}
\vspace{-2mm}
\caption{\label{table:mip360} Performance comparisons of sparse-view synthesis on Mip-NeRF360 dataset~\cite{barron22mip360}. The best, second-best, and third-best entries are marked in red, orange, and yellow, respectively.}
\vspace{-5mm}
\end{table*}

The performance comparison on the Mip-NeRF360 dataset is also shown in Table~\ref{table:mip360}.
It is noteworthy that our approach attains 23.92 (PSNR), 0.755 (SSIM), and 0.242 (LPIPS), which significantly outperforms the state-of-the-art methods by a meaningful margin.
Finally, we also evaluate the rendering performance on the Blender dataset, and the corresponding result is shown in Table~\ref{table:blender}. 
The proposed method achieves the highest PSNR of 25.42 while the performance for other two metrics are slightly dropped compared to the best scores. Even though the proposed method always shows the best performance for all the metrics, it is still effective and competitive for rendering with sparse view inputs without requiring any additional module or algorithm.

Besides, we compare the proposed method with feed-forward 3DGS methods, such as pixelSplat~\cite{charatan23pixelsplat}, MVSplat~\cite{chen2024mvsplat}, and FreeSplat~\cite{wang2025freesplat}, on the Replica dataset (see Table~\ref{table:comparison_feedforward}).
While these pre-trained feed-forward models can provide faster inference, our method still demonstrates high visual quality under sparse-view inputs.

\begin{figure}
\centering
\centerline{\includegraphics[width=1.0\columnwidth]{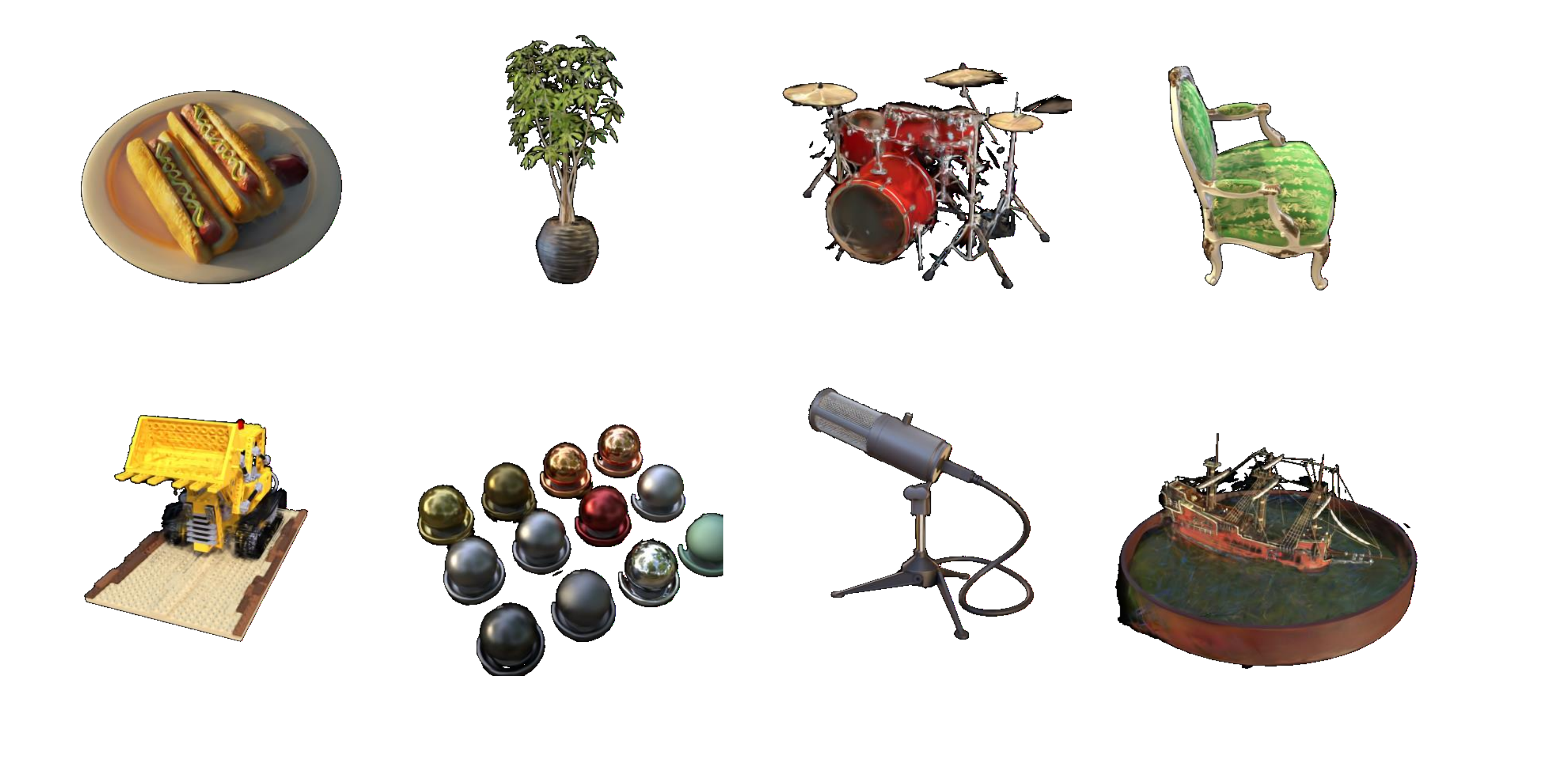}}
\vspace{-5mm}
\caption{\label{fig:7}
Results of novel view rendering on the Blender dataset~\cite{mildenhall20NeRF} by the proposed method.}
\vspace{-3mm}
\end{figure}

\noindent\textbf{Qualitative evaluation.}
Furthermore, the qualitative comparison of the proposed method with FSGS~\cite{zhu24FSGS}, CoR-GS~\cite{zhang24CoR-GS}, and 3DGS on the LLFF dataset is presented in Fig.~\ref{fig:6}. The results demonstrate the effectiveness of our approach in forward-facing scenes, particularly in achieving high precision and artifact-free renderings compared to baseline methods.
In the first row of Fig.~\ref{fig:6}, our method consistently outperforms other approaches in rendering forward-facing scenes with greater precision.
While other methods exhibit visible inaccuracies in capturing fine details and maintaining structural coherence, our approach achieves superior fidelity, preserving intricate scene features and producing more realistic outputs.
In the second row, the robustness of the proposed method in mitigating overfitting is further demonstrated.
FSGS and CoR-GS exhibit prominent artifacts that compromise the quality and realism of the renderings.
In contrast, our method effectively avoids these artifacts, maintaining clean and accurate reconstructions even in challenging regions.
Rendering results from novel views by the proposed method are also shown in Fig.~\ref{fig:7}.
These qualitative results reinforce the effectiveness of our method in addressing overfitting and ensuring robust performance across different datasets and scene types, showcasing its versatility and reliability in a wide range of rendering tasks.

\subsection{Limitations and Future Work}
While the proposed method demonstrates significant improvements in mitigating overfitting and enhancing rendering quality in sparse-view 3D Gaussian splatting (3DGS), there are certain limitations that require further investigation.
These limitations highlight opportunities for future research and practical enhancements.  The reliance on hyperparameters, such as the scaling factor \(\gamma\) for the dropping rate \(r\), introduces sensitivity to dataset-specific tuning.
While \(\gamma\) plays a crucial role in progressively adjusting the dropping rate to mitigate overfitting, its optimal value can vary depending on the dataset and task. Future work could explore adaptive mechanisms to dynamically adjust \(\gamma\) during training, reducing the need for manual fine-tuning and improving generalization across diverse datasets.

\begin{table}[t]
\scriptsize
\centering
\resizebox{1.0\columnwidth}{!}{
\begin{tabular}{c l | c c c} 
\toprule
\multicolumn{2}{c}{Methods} & \multicolumn{1}{c}{PSNR($\uparrow$)} & \multicolumn{1}{c}{SSIM($\uparrow$)} & \multicolumn{1}{c}{LPIPS ($\downarrow$)}\\ 
\midrule
\multirow{5}{*}[0ex]{\centering\rotatebox{90}{NeRF-based}} 
& Mip-NeRF~\cite{barron21mip} & 20.89 &  0.830 & 0.168\\
& DietNeRF~\cite{jain21dietnerf} & 22.50 & 0.823 &  0.124\\ 
& RegNeRF~\cite{niemeyer2022regnerf} & 23.86 &  0.852 & 0.105 \\
& FreeNeRF~\cite{yang23freenerf} & 24.26 & 0.883 & 0.098\\
& SparseNeRF~\cite{wang23sparsenerf} & 24.04 & 0.876 & 0.113\\
\midrule
\multirow{5}{*}[0ex]{\centering\rotatebox{90}{3DGS-based}} 
& 3DGS~\cite{kerbl233DGS} & 21.56 & 0.847 & 0.130\\
& DNGaussian~\cite{li24Dngaussian} & 24.31 & 0.886 & 0.088\cellcolor{orange!25}\\
& FSGS~\cite{zhu24FSGS} & 24.64\cellcolor{orange!25} & 0.895 \cellcolor{orange!25} & 0.095\\
& CoR-GS~\cite{zhang24CoR-GS} & 24.43\cellcolor{yellow!25} & 0.896\cellcolor{red!25} & 0.084\cellcolor{red!25} \\
& Ours & 25.42\cellcolor{red!25} & 0.888\cellcolor{yellow!25} & 0.089\cellcolor{yellow!25}\\ 
\bottomrule
\end{tabular}
}
\vspace{-2mm}
\caption{\label{table:blender} Performance comparisons of sparse-view synthesis on Blender dataset~\cite{mildenhall20NeRF}. The best, second-best, and third-best entries are marked in red, orange, and yellow, respectively.}
\vspace{-3mm}
\end{table}

\begin{table}[t]
\small
\centering
\resizebox{1.0\columnwidth}{!}{
\begin{tabular}{l | l | c c c}
\toprule
\multicolumn{1}{c}{Methods} & \multicolumn{1}{c}{Type} & \multicolumn{1}{c}{PSNR ($\uparrow$)} & \multicolumn{1}{c}{SSIM ($\uparrow$)} & \multicolumn{1}{c}{LPIPS ($\downarrow$)}\\ 
\midrule
pixelSplat~\cite{charatan23pixelsplat} & Feed-forward & 26.24\cellcolor{yellow!25} & 0.829 & 0.229\\
MVSplat~\cite{chen2024mvsplat} & Feed-forward & 26.16 & 0.840 \cellcolor{yellow!25}& 0.173\cellcolor{orange!25}\\
FreeSplat~\cite{wang2025freesplat} & Feed-forward & 26.98\cellcolor{orange!25} & 0.848\cellcolor{orange!25} & 0.171\cellcolor{red!25}\\
Ours & Optimization & 27.76\cellcolor{red!25} & 0.866\cellcolor{red!25} & 0.179\cellcolor{yellow!25}\\
\bottomrule
\end{tabular}
}
\vspace{-2mm}
\caption{\label{table:comparison_feedforward}Performance comparison with feed-forward methods on the Replica dataset (2-view)~\cite{straub2019replica}. The best, second-best, and third-best entries are marked in red, orange, and yellow, respectively.}
\vspace{-6mm}
\end{table}
\subsection{Ablation Study}
In this subsection, we check the effect of the change in dropping rates and the progressive adjustment strategy.
The performance for all the experiments in this subsection is evaluated on the LLFF~\cite{mildenhall19llff} dataset with the 3-view setting.
Note that the scale factor of the dropping rate becomes the dropping rate when the progressive adjustment strategy is not used.
Specifically, compared to the use of the fixed dropping rate, our progressive adjustment strategy efficiently improves the rendering performance regardless of the value of the dropping rate as shown in Table~\ref{table:ablation_study}.
For the fixed dropping rates, the performance degradation was observed as the ratio of dropping Gaussians increased from 0.1 to 0.3, which indicates the adverse effect of overly aggressive Gaussian removal. One interesting point is that this reversal highlights the effectiveness of progressively increasing the dropping rate in alleviating the overfitting problem, particularly during the later stage of training. The best performance is achieved when using 0.2 for the scale factor of the dropping rate, thus \(\gamma=0.2\) is our default setting.
These findings demonstrate the utility of our progressive adjustment of the dropping rate for reducing overfitting while improving the rendering quality.
Furthermore, we observe that randomly dropping Gaussian primitives is more effective than selectively dropping them.
Since selective approaches rely on metrics like gradient magnitude and distance, they have a potential risk repeatedly discarding Gaussians that are critical for reconstructing scenes under sparse-view conditions.
The corresponding result is shown in Table~\ref{table:ablation_study2}.
In the domain of optimization, L1 regularization is commonly used to prune elements that exhibit a lower correlation with the objective function.
However, this approach may permanently remove Gaussians, whereas our method only temporarily deactivates Gaussians.
That is, strongly correlated Gaussians can re-engage during later iterations.
This property efficiently improves the performance compared to L1 regularization as shown in Table~\ref{table:ablation_study3}.

\begin{table}[t]
\scriptsize
\centering
\resizebox{\columnwidth}{!}{ 
\begin{tabular}{c c | c c c}
\toprule
\multicolumn{1}{c}{Scaling factor \((\gamma)\)} & \multicolumn{1}{c}{Progressive} & \multicolumn{1}{c}{PSNR ($\uparrow$)} & \multicolumn{1}{c}{SSIM ($\uparrow$)} & \multicolumn{1}{c}{LPIPS ($\downarrow$)}\\
\midrule
- & - & 19.22 & 0.649 & 0.229\\
\midrule
0.1 &\ding{55}  & 20.28 & 0.701 & 0.209\cellcolor{yellow!25}\\
0.2 &\ding{55}  & 20.16 & 0.700 & 0.216\\
0.3 & \ding{55} & 20.15 & 0.691 & 0.223\\
0.1 & \checkmark & 20.29\cellcolor{yellow!25} & 0.702\cellcolor{yellow!25} & 0.206\cellcolor{orange!25}\\
0.2 & \checkmark & 20.56\cellcolor{red!25} & 0.708 \cellcolor{red!25}& 0.207\cellcolor{red!25}\\
0.3 & \checkmark & 20.48\cellcolor{orange!25} & 0.707 \cellcolor{orange!25}& 0.212\\
\bottomrule
\end{tabular}
}
\vspace{-2mm}
\caption{\label{table:ablation_study}Ablation study on the LLFF dataset (3-view). The best, second-best, and third-best entries are marked in red, orange, and yellow, respectively.}
\vspace{-3mm}
\end{table}

\begin{table}[t]
\scriptsize
\centering
\resizebox{1.0\columnwidth}{!}{
\begin{tabular}{c | c | c c c}
\toprule
\multicolumn{1}{c}{Methods} & \multicolumn{1}{c}{Metrics} & \multicolumn{1}{c}{PSNR ($\uparrow$)} & \multicolumn{1}{c}{SSIM ($\uparrow$)} & \multicolumn{1}{c}{LPIPS ($\downarrow$)}\\ 
\midrule
Random & - & 20.76\cellcolor{red!25} & 0.713\cellcolor{red!25} & 0.200\cellcolor{red!25}\\
Selective & Gradient & 19.62\cellcolor{orange!25} & 0.682\cellcolor{orange!25} & 0.214\cellcolor{orange!25}\\
Selective & Distance & 17.49\cellcolor{yellow!25} & 0.582\cellcolor{yellow!25} & 0.273\cellcolor{yellow!25}\\
\bottomrule
\end{tabular}
}
\vspace{-2mm} 
\caption{\label{table:ablation_study2}Ablation study according to dropping schemes. The best, second-best, and third-best entries are marked in red, orange, and yellow, respectively.}
\vspace{-3mm} 
\end{table}

\begin{table}[t]
\scriptsize
\centering
\resizebox{1.0\columnwidth}{!}{
\begin{tabular}{c | c | c c c}
\toprule
\multicolumn{1}{c}{Methods} & \multicolumn{1}{c}{Metrics} & \multicolumn{1}{c}{PSNR ($\uparrow$)} & \multicolumn{1}{c}{SSIM ($\uparrow$)} & \multicolumn{1}{c}{LPIPS ($\downarrow$)}\\ 
\midrule
Dropping & - & 20.76\cellcolor{red!25} & 0.713\cellcolor{red!25} & 0.200\cellcolor{red!25}\\
L1 reg.& Gradient & 19.97\cellcolor{orange!25} & 0.690\cellcolor{orange!25} & 0.211\cellcolor{orange!25}\\
L1 reg.& Distance & 19.73\cellcolor{yellow!25} & 0.681\cellcolor{yellow!25} & 0.213\cellcolor{yellow!25}\\
\bottomrule
\end{tabular}
}
\vspace{-2mm} 
\caption{\label{table:ablation_study3}Ablation study according to regularization schemes. The best, second-best, and third-best entries are marked in red, orange, and yellow, respectively.}
\vspace{-5mm} 
\end{table}

\vspace{-2mm}
\section{Conclusion}
\vspace{-2mm}
In this paper, we introduce a simple yet powerful\linebreak method for sparse-view 3DGS.
The key idea of the proposed method, so-called DropGaussian, is to mitigate\linebreak the overfitting problem by randomly removing 3D Gaussians during the training process.
By simple dropping\linebreak operations, Gaussians, which are far from the camera in sparse-view conditions, can have larger gradients while\linebreak improving their visibility.
This is fairly desirable to\linebreak mitigate the overfitting problem in rendering with sparse\linebreak input views. 
Moreover, we propose to progressively\linebreak apply our DropGaussian to the training process for fur-\linebreak ther enhancing the visual quality of the rendering result.\linebreak

\vspace{-2mm}
\noindent\textbf{Acknowledgments.} This work was supported by the National Research Foundation of Korea (NRF) grant funded by the Korea government (MSIT) (RS-2023-NR076462) and Institute of Information Communications Technology Planning Evaluation (IITP) grant funded by the Korea government (MSIT) (No. 2018-0-00207, RS-2018-II180207, Immersive Media Research Laboratory).

{
    \small
    \bibliographystyle{ieeenat_fullname}
    \bibliography{main}
}


\end{document}